\documentclass[11pt,a4paper]{article}
\usepackage[hyperref]{emnlp-ijcnlp-2019}
\usepackage{times}
\usepackage{latexsym}
\usepackage{enumitem}
\usepackage{url}
\usepackage{tabularx}

\RequirePackage[textsize=scriptsize]{todonotes}
\usepackage{bm,amsmath,amsfonts,amssymb}
\usepackage{soul}
\usepackage[noend,ruled,vlined]{algorithm}
\usepackage[noend]{algpseudocode}
\usepackage{float}
\usepackage{multicol}

\usepackage[normalem]{ulem}
\usepackage{etoolbox}
\usepackage{tikz}
\usetikzlibrary{tikzmark}
\usetikzlibrary{calc}

\errorcontextlines\maxdimen
\newcommand{\ALGtikzmarkcolor}{black}
\newcommand{\ALGtikzmarkextraindent}{4pt}
\newcommand{\ALGtikzmarkverticaloffsetstart}{-.5ex}
\newcommand{\ALGtikzmarkverticaloffsetend}{-.5ex}
\makeatletter
\newcounter{ALG@tikzmark@tempcnta}

\newcommand\ALG@tikzmark@start{%
    \global\let\ALG@tikzmark@last\ALG@tikzmark@starttext%
    \expandafter\edef\csname ALG@tikzmark@\theALG@nested\endcsname{\theALG@tikzmark@tempcnta}%
    \tikzmark{ALG@tikzmark@start@\csname ALG@tikzmark@\theALG@nested\endcsname}%
    \addtocounter{ALG@tikzmark@tempcnta}{1}%
}

\def\ALG@tikzmark@starttext{start}
\newcommand\ALG@tikzmark@end{%
    \ifx\ALG@tikzmark@last\ALG@tikzmark@starttext
    \else
        \tikzmark{ALG@tikzmark@end@\csname ALG@tikzmark@\theALG@nested\endcsname}%
        \tikz[overlay,remember picture] \draw[\ALGtikzmarkcolor] let \p{S}=($(pic cs:ALG@tikzmark@start@\csname ALG@tikzmark@\theALG@nested\endcsname)+(\ALGtikzmarkextraindent,\ALGtikzmarkverticaloffsetstart)$), \p{E}=($(pic cs:ALG@tikzmark@end@\csname ALG@tikzmark@\theALG@nested\endcsname)+(\ALGtikzmarkextraindent,\ALGtikzmarkverticaloffsetend)$) in (\x{S},\y{S})--(\x{S},\y{E});%
    \fi
    \gdef\ALG@tikzmark@last{end}%
}

\apptocmd{\ALG@beginblock}{\ALG@tikzmark@start}{}{\errmessage{failed to patch}}
\pretocmd{\ALG@endblock}{\ALG@tikzmark@end}{}{\errmessage{failed to patch}}
\makeatother

\aclfinalcopy 

\newcommand{\vek}[1]{{\bf {#1}}}
\newcommand{\vx}{{\vek{x}}}
\newcommand{\vc}{{\vek{c}}}
\newcommand{\ve}{{\vek{e}}}
\newcommand{\pos}{{p}}
\newcommand{\vpos}{{\vek{\pos}}}
\newcommand{\xo}{y}

\newcommand{\vs}{{\vek{s}}}
\newcommand{\vh}{{\vek{h}}}
\newcommand{\vxo}{{\vek{\xo}}}

\newcommand{\vocab}{{\Sigma}}
\newcommand{\cpy}{\mbox{\sc c}}
\newcommand{\app}{\mbox{\sc a}}
\newcommand{\ins}{\mbox{\sc i}}
\newcommand{\subs}{\mbox{\sc s}}
\newcommand{\delete}{\mbox{\sc d}}
\newcommand{\xform}{{\mbox{\sc t}}}
\newcommand{\replace}{\mbox{\sc r}}
\newcommand{\vtheta}{{\bm{\theta}}}
\newcommand{\editF}{{\text{Seq2Edits}}}

\newcommand{\addsuffix}{\mbox{\sc AddSuffix}}

\newcommand{\transform}[2]{\sc Change-#1-to-#2}
\newcommand{\vr}{{\vek{r}}}
\newcommand{\vi}{{\vek{a}}}
\newcommand{\vg}{{\vek{h}}}
\newcommand{\vu}{{\vek{h}}}
\newcommand{\cE}{{\cal E}}
\newcommand{\cL}{{\mathcal L}}
\newcommand{\maskE}{{\mbox{\sc m}}}
\newcommand{\maskT}{{\mbox{\sc mask}}}
\newcommand\numberthis{\addtocounter{equation}{1}\tag{\theequation}}

\title{Parallel Iterative Edit Models for Local Sequence Transduction}

\author{Abhijeet Awasthi\thanks{\enspace Correspondence to: {\tt awasthi@cse.iitb.ac.in}}
  ,
  Sunita Sarawagi
  ,
  Rasna Goyal 
  ,
  Sabyasachi Ghosh 
  ,
  Vihari Piratla \\
  Department of Computer Science and Engineering, IIT Bombay
  }

\begin{document}
\maketitle
\begin{abstract}
We present a Parallel Iterative Edit (PIE) model for the problem of local sequence transduction arising in tasks like Grammatical error correction (GEC). Recent  approaches are based on the popular encoder-decoder (ED) model for sequence to sequence learning. The ED model auto-regressively captures full dependency among output tokens but is slow due to sequential decoding. The PIE model does parallel decoding, giving up the advantage of modelling full dependency in the output, yet it achieves accuracy competitive with the ED model for four reasons: 1.~predicting edits instead of tokens, 2.~labeling sequences instead of generating sequences, 3.~iteratively refining predictions to capture dependencies, and 4.~factorizing logits over edits and their token argument to harness pre-trained language models like BERT. Experiments on tasks spanning GEC, OCR correction and spell correction demonstrate that the PIE model is an accurate and significantly faster alternative for local sequence transduction. The code and pre-trained models for GEC are available at \url{https://github.com/awasthiabhijeet/PIE}.
\end{abstract}

\section{Introduction}
\label{submission}
In local sequence transduction (LST) an input sequence $x_1,\ldots,x_n$ needs to be mapped to an output sequence $\xo_1,\ldots,\xo_m$ where the $\vx$ and $\vxo$ sequences differ only in a few positions, $m$ is close to $n$, and $x_i, \xo_j$ come from the same vocabulary $\vocab$. An important application of local sequence transduction that we focus on in this paper is Grammatical error correction (GEC). We contrast local transduction with more general sequence transduction tasks like translation and paraphrasing which might entail different input-output vocabulary and non-local alignments. The general sequence transduction task is cast as sequence to sequence (seq2seq) learning and modeled popularly using an attentional encoder-decoder (ED) model. The ED model  auto-regressively produces each token $y_t$ in the output sequence conditioned on all previous tokens $y_1,\ldots,y_{t-1}$. Owing to the remarkable success of this model in challenging tasks like translation, almost all state-of-the-art neural models for GEC use it ~\cite{zhao2019improving,lichtarge2019,ge2018reaching,chollampatt2018neural,junczys2018approaching}. 

We take a fresh look at local sequence transduction tasks and present a new parallel-iterative-edit (PIE) architecture.  Unlike the prevalent ED model that is constrained to sequentially generating the tokens in the output, the PIE model generates the output in parallel,  thereby substantially reducing the latency of sequential decoding on long inputs.
However, matching the accuracy of existing ED models without the luxury of conditional generation is highly challenging. Recently, parallel models have also been explored in tasks like translation~\cite{Stern2018,Lee2018,gu2018,kaiser18a} and speech synthesis~\cite{oord18a}, but their accuracy is significantly lower than corresponding ED models. The PIE model incorporates the following four ideas to achieve comparable accuracy on tasks like GEC in spite of parallel decoding.

\noindent{\bf 1.~Output edits instead of tokens:}
First, instead of outputting tokens from a large vocabulary, we output edits such as copy, appends, deletes, replacements, and case-changes which generalize better across tokens and yield a much smaller vocabulary. Suppose in GEC we have an input sentence: {\tt fowler fed dog}. Existing seq2seq learning approaches would need to output the four tokens {\tt Fowler, fed, the, dog} from a word vocabulary whereas we would predict the edits \{{\tt Capitalize token 1, Append(the) to token 2, Copy token 3}\}. 

\noindent{\bf 2.~Sequence labeling instead of sequence generation:}
Second we perform in-place edits on the source tokens and formulate local sequence transduction as {\em labeling} the input tokens with edit commands, rather than solving the much harder whole \emph{sequence generation} task involving a separate decoder and attention module.
Since input and output lengths are different in general such formulation is non-trivial, particularly due to edits that insert words.  We create special compounds edits that merge token inserts with preceding edits that yield higher accuracy than earlier methods of independently predicting inserts~\cite{Ribeiro2018,Lee2018}.  

\noindent{\bf 3.~Iterative refinement:}
Third, we increase the inference capacity of the parallel model by  iteratively inputting the model's own output for further refinement.  This handles dependencies implicitly, in a way reminiscent of Iterative Conditional Modes (ICM) fitting in graphical model inference~\cite{daphneBook}. \citet{lichtarge2019} and \citet{ge2018reaching} also refine iteratively but with ED models.

\noindent{\bf 4.~Factorize pre-trained bidirectional LMs:}
Finally, we adapt recent pre-trained bidirectional models like BERT~\cite{devlin2018bert} by factorizing the logit layer over edit commands and their token argument.  Existing GEC systems typically rely on conventional forward directional LM to pretrain their decoder, whereas we show how to use a bi-directional LM in the encoder, and that too to predict edits.\\  

\noindent{Novel contributions of our work are as follows:}
\begin{itemize}[leftmargin=*]
\item Recognizing GEC as a local sequence transduction (LST) problem, rather than machine translation. We then cast LST as a fast non-autoregressive, sequence labeling model as against existing auto-regressive encoder-decoder model.
\item Our method of reducing LST to non-autoregressive sequence labeling has many novel elements: outputting edit operations instead of tokens, append operations instead of insertions in the edit space, and replacements along with custom transformations.
\item We show how to effectively harness a pre-trained language model like BERT using our factorized logit architecture with edit-specific attention masks.
\item The parallel inference in PIE is 5 to 15 times faster than a competitive ED based GEC model like \cite{lichtarge2019} which performs sequential decoding using beam-search. PIE also attains close to state of the art performance on standard GEC datasets. On two other local transduction tasks, viz., OCR and spell corrections the PIE model is fast and accurate w.r.t. other existing models developed specifically for local sequence transduction.
\end{itemize}

\section{Our Method} \label{sec:method}
We assume a fully supervised training setup where we are given a parallel dataset of incorrect, correct sequence pairs: $D=\{(\vx^i,\vxo^i):i=1 \ldots N\}$, and an optional  large corpus of unmatched correct sequences $\cL=\{\tilde{\vxo}^1,\ldots, \tilde{\vxo}^U\}$. 
In GEC, this could be a corpus of grammatically correct sentences used to pre-train a language model.  

\paragraph{Background: existing ED model} Existing seq2seq ED models factorize $\Pr(\vxo|\vx)$ to capture the full dependency between a $\xo_t$ and all previous $\vxo_{<t}=\xo_1,\ldots,\xo_{t-1}$ as $\prod_{t=1}^m\Pr(\xo_t|\vxo_{<t},\vx)$.  An encoder converts input tokens $x_1,\ldots,x_n$ to contextual states $\vh_1,\ldots,\vh_n$ and a decoder summarizes $\vxo_{<t}$ to a state $\vs_t$. An attention distribution over contextual states computed from $\vs_t$ determines the relevant input context $\vc_t$ and the output token distribution is calculated as $\Pr(\xo_t|\vxo_{<t},\vx) = \Pr(\xo_t|\vc_t,\vs_t)$. Decoding is done sequentially using beam-search. When a correct sequence corpus $\cL$ is available, the decoder is pre-trained  on a next-token prediction loss and/or a trained LM is used to re-rank the beam-search outputs~\cite{zhao2019improving, chollampatt2018mlconv, junczys2018approaching}.

\subsection{Overview of the PIE model}
We move from generating tokens in the output sequence $\vxo$ using a separate decoder, to labelling the input sequence $x_1,\ldots,x_n$ with edits $e_1,\ldots,e_n$.
For this we need to design a function \editF\ that takes as input an $(\vx,\vxo)$
pair in $D$ and outputs a sequence $\ve $ of edits from an edit space
$\cE$ where $\ve $ is of the same length as $\vx$ in spite of $\vx$
and $\vxo$ being of different lengths.  In Section~\ref{sec:editseq}
we show how we design such a function.

\paragraph{Training:}
We invoke \editF\ on $D$ and learn the parameters of a probabilistic model $\Pr(\ve|\vx, \theta)$ to assign a distribution over the edit labels on tokens of the input sequence. In Section~\ref{sec:arch} we describe the PIE architecture in more detail.
The correct corpus $\cL$ , when available, is used to pre-train the encoder to predict an arbitrarily masked token $y_t$ in a sequence $\vxo$ in $\cL$, much like in BERT.  Unlike in existing seq2seq systems where $\cL$ is used to pre-train the decoder that only captures forward dependencies, in our pre-training the predicted token $y_t$ is dependent on both forward and backward contexts. This is particularly useful for GEC-type tasks where future context $\xo_{t+1}\ldots \xo_m$ can be approximated by $x_{t+1},\ldots,x_{n}$. 

\paragraph{Inference: } Given an input $\vx$, the trained model predicts the edit distribution for each input token independent of others, that is $\Pr(\ve|\vx, \theta)=\prod_{t=1}^n \Pr(e_t|\vx, t, \theta)$,  and thus does not entail the latency of sequential token generation of the ED model. We output the most probable edits $\hat{\ve}= \text{argmax}_\ve \Pr(\ve|\vx, \theta)$.
Edits are designed so that we can easily get the edited sequence $\hat{\vxo}$ after {\em applying} $\hat{\ve}$ on $\vx$. $\hat{\vxo}$ is further refined by iteratively applying the model on the generated outputs $\hat{\vxo}$ until we get a sequence identical to one of the previous sequences upto a maximum number of iterations $I$.

\subsection{The  \editF\ Function}
\label{sec:editseq}
Given a $\vx=x_1,\ldots,x_n$ and $\vxo=\xo_1,\ldots,\xo_m$ where $m$ may not be equal to $n$,
our goal is to obtain a sequence of edit operations $\ve = (e_1,\ldots,e_n): e_i \in \cE$ such that applying edit $e_i$ on the input token $x_i$ at each position $i$ reconstructs the output sequence $\vxo$.
Invoking an off-the-shelf edit distance algorithm between $\vx$ and $\vxo$ can give us a sequence of copy, delete, replace, and insert operations of arbitrary length.  The main difficulty is converting the insert operations into in-place edits at each $x_i$. Other parallel models~\cite{Ribeiro2018,Lee2018} have used methods like predicting insertion slots in a pre-processing step, or predicting zero or more tokens in-between any two tokens in $\vx$.  We will see in Section~\ref{sec:ablation} that these options do not perform well.  Hence we design an alternative edit space $\cE$ that merges inserts with preceding edit operations creating compound append or replace operations.
Further, we create a  dictionary $\Sigma_a$ of common q-gram insertions or replacements observed in the training data.\\
Our edit space (${\mathcal E}$) comprises of copy (\cpy) $x_i$, delete (\delete) $x_i$, append (\app) a  q-gram $w \in \Sigma_a$ after copying $x_i$, replace (\replace) $x_i$ with a q-gram $w \in \Sigma_a$. For GEC, we additionally use transformations denoted as $\xform_1,\ldots,\xform_k$ which perform word-inflection (e.g. {\tt arrive} to {\tt arrival}).
The space of all edits is thus:
\begin{multline}
\label{eq:edits}
    {\mathcal E} = \{
\cpy,
\delete,
\xform_1,\ldots,\xform_k\} \\
\cup \{\app(w): w \in \Sigma_a\} \\
\cup \{\replace(w): w \in \Sigma_a\} 
\end{multline}

\begin{table}[h]
    \setlength\tabcolsep{1.0pt}
\begin{small}
\begin{tabular}{l|cccccccccc} \hline
 &  \multicolumn{10}{c}{Example 1}  \\ \hline
     $\vx$ &  \multicolumn{10}{l}{~[ Bolt can have run race ]} 
      
\\ 
      $\vxo$ & \multicolumn{10}{l}{  ~[ Bolt could have run the race ]} 
      
\\ 

{\tt diff} & \multicolumn{10}{l}{~(\cpy,{\tt [}) (\cpy,{\tt Bolt}) (\delete,{\tt can}) (\ins,{\tt can},{\tt could}) (\cpy,{\tt have})}
\\
 & \multicolumn{10}{l}{~ (\cpy,{\tt run}) (\ins,{\tt run},{\tt the}) (\cpy,{\tt race}) (\cpy,{\tt]})}
  \\ \hline
$\ve$ &\cpy & \cpy & \replace({\tt could})& \cpy & \app({\tt the}) & \cpy  & \cpy & \hspace{.63cm}

\\ 
& $\uparrow$ & $\uparrow$ & $\uparrow$& $\uparrow$& $\uparrow$& $\uparrow$& $\uparrow$&
\\
& [ & Bolt & can & have & run & race & ] & & \\
\end{tabular}

\begin{tabular}{l|ccccccccc}\hline 
&  \multicolumn{9}{c}{Example 2}  \\ \hline
     $\vx$ &  \multicolumn{9}{l}{~[ He still won race ! ]} 
     
\\ 
      $\vxo$ & \multicolumn{9}{l}{  ~[ However , he still won ! ]} 
       
\\ 
{\tt diff} & \multicolumn{9}{l}{~(\cpy,{\tt [}) (\ins,{\tt [},{\tt However,}) (\delete,{\tt He}) (\ins,{\tt He},{\tt he}) (\cpy,{\tt still}) }
\\
 & \multicolumn{9}{l}{~ (\cpy,{\tt won}) (\delete,\tt{-race}) (\cpy,\tt{!}) (\cpy,\tt{]})}

  \\ \hline
$\ve$ &~\app({\tt{However,}}) & \xform\_\text{case} & \cpy& \cpy & \delete & \cpy  & \cpy & \hspace{1cm}  & 
\\ 
& $\uparrow$ & $\uparrow$ & $\uparrow$& $\uparrow$& $\uparrow$& $\uparrow$& $\uparrow$ &  &   
\\
& [ & He & still & won & race & ! & ] &  & 

\\ 
\hline 
\end{tabular}
\caption{Two example sentence pairs converted into respective edit sequences.}
\label{tab:example}
\end{small}
\end{table}
\setlength\tabcolsep{5.0pt}

\begin{figure}
\begin{small}
\begin{algorithmic}
\Require{$\vx = (x_1,\ldots,x_n)$, $\vxo=(\xo_1,\ldots,\xo_m)$} and  $\mathcal{T}:$ List of Transformations
\State diffs $\gets \Call{Levenshtein-Dist}{\vx,\vxo}$ with modified cost.

\State diffs $\gets$ In diffs break substitutions, merge inserts into q-grams
\State $\Sigma_a \gets$ $M$ most common inserts in training data
\State $\ve \gets \Call{EmptyArray}{n}$
\For {$t \gets 1$ to \Call{length}{\text{diffs}}}
    \If{$\text{diffs}[t]=(\cpy, x_i)$ or $(\delete, x_i)$}
        \State $e_i \gets \text{diffs}[t].op$
    \ElsIf{$\text{diffs}[t]=(\ins,x_i,w)$}
       \If{$e_i=\delete$} 
            \If{$(x_i,w)$ match transformation T $\in \mathcal{T}$}
                \State $e_i \gets \text{T}$
            \Else
                \State $e_i \gets \replace(w)~ \textbf{if}~ w \in \Sigma_a~ \text{else}~ \cpy $  
            \EndIf
       \ElsIf{$e_i=\cpy$}
            \State $e_i \gets \app(w)~ \textbf{if}~ w \in \Sigma_a~ \text{else}~ \cpy$
        \EndIf
    \EndIf
\EndFor
\Return{$\ve$}
\end{algorithmic}
\caption{The \editF\ function used to convert a sequence pair $\vx,\vxo$ into in-place edits.}
\label{algo2:generate_edits}
\end{small}
\end{figure}

We present our algorithm for converting a sequence $\vx$ and $\vxo$ into in-place
edits on $\vx$ using the above edit space in 
Figure~\ref{algo2:generate_edits}. Table~\ref{tab:example} gives
examples of converting ($\vx$, $\vxo$) pairs to edit sequences.
We first invoke the Levenshtein distance algorithm
\citep{levenshtein1966binary} to obtain {\tt diff} between $\vx$ and
$\vxo$ with delete and insert cost as 1 as usual, but with a modified
substitution cost to favor matching of related words. 
We detail this modified cost in the Appendix and show
an example of how this modification leads to more sensible
edits.  The {\tt diff} is post-processed to convert substitutions
into deletes followed by inserts, and consecutive inserts are merged
into a q-gram. We then create a dictionary $\Sigma_a$ of the $M$ most frequent
q-gram inserts in the training set.
Thereafter, we scan the {\tt diff} left to right: a copy at $x_i$ makes $e_i$ = \cpy, a delete at $x_i$ makes $e_i=\delete$, an insert $w$ at $x_i$ and a $e_i=\cpy$ flips the $e_i$ into a $e_i=\app(w)$ if $w$ is in $\Sigma_a$, else it is dropped, an insert $w$ at $x_i$ and a $e_i=\delete$ flips the $e_i$ 
into a $e_i=\xform(w)$ if a match found, else a replace $\replace(w)$ if $w$ is in $\Sigma_a$, else it is dropped.

The above algorithm does not guarantee that when $\ve$ is applied on
$\vx$ we will recover $\vxo$ for all sequences in the training
data.  This is because we limit $\Sigma_a$  to include only
the $M$ most frequently inserted q-grams.  For local sequence transduction tasks,
we expect a long chain of consecutive inserts to be rare, hence our
experiments were performed with $q=2$.
For example, in NUCLE dataset \cite{ng2014conll} which has roughly 57.1K sentences, less than 3\%  sentences have three or more consecutive inserts.

\subsection{The Parallel Edit Prediction Model}
\label{sec:arch}
We next describe our model for predicting edits $\ve: e_1,\ldots,e_n$ on an input sequence $\vx: x_1,\ldots,x_n$.  We use a bidirectional encoder to provide a contextual encoding of each $x_i$. This can either be multiple layers of bidirectional RNNs, CNNs or deep bidirectional transformers.  We adopt the deep bidirectional transformer architecture since it encodes the input in parallel.   
We pre-train the model using $\cL$ much like in
BERT pre-training recently proposed for language modeling\cite{devlin2018bert}. We first give an overview of BERT and then describe our model.

\paragraph{Background: BERT}
The input is token embedding $\vx_i$ and positional embedding $\vpos_i$ for each token $x_i$ in the sequence $\vx$.  Call these jointly as  $\vh^0_i = [\vx_i,\vpos_i]$.  Each layer $\ell$ produces a  $\vh^\ell_i$ at each position $i$ as a function of $\vh^{\ell-1}_i$ and self-attention over all $\vh^{\ell-1}_j$, $j \in [1,n]$. The BERT model is pre-trained by masking out a small fraction of the input tokens by a special \maskT,  
and predicting the masked word from its bidirectional context captured as the last layer output: $\vh_1,\ldots,\vh_n$.

\subsubsection{A Default use of BERT}
\label{sec:default}
Since we have cast our task as a sequence labeling task, a default output layer would be to compute $\Pr(e_i|\vx)$ as a softmax over the space of edits ${\cal E}$ from each $\vh_i$. If $W_e$ denotes the softmax parameter for edit $e$, we get:   
\begin{equation}
\label{eq:nonfac}
    \Pr(e_i=e|\vx) = \text{softmax}(W_{e}^{\intercal}\vh_i)
\end{equation}
The softmax parameters, $W_e$ in Equation~\ref{eq:nonfac}, have to be trained from scratch. We propose a method to exploit token embeddings of the pre-trained language model to warm-start the training of edits like appends and replaces which are associated with a token argument.
Furthermore, for appends and replaces, we provide a new method of computing the hidden layer output 
via alternative input positional embeddings and  self-attention. We do so without  introducing any new parameters in the hidden layers of BERT.

\begin{figure}
    \centering
    \includegraphics[width=1.0\hsize]{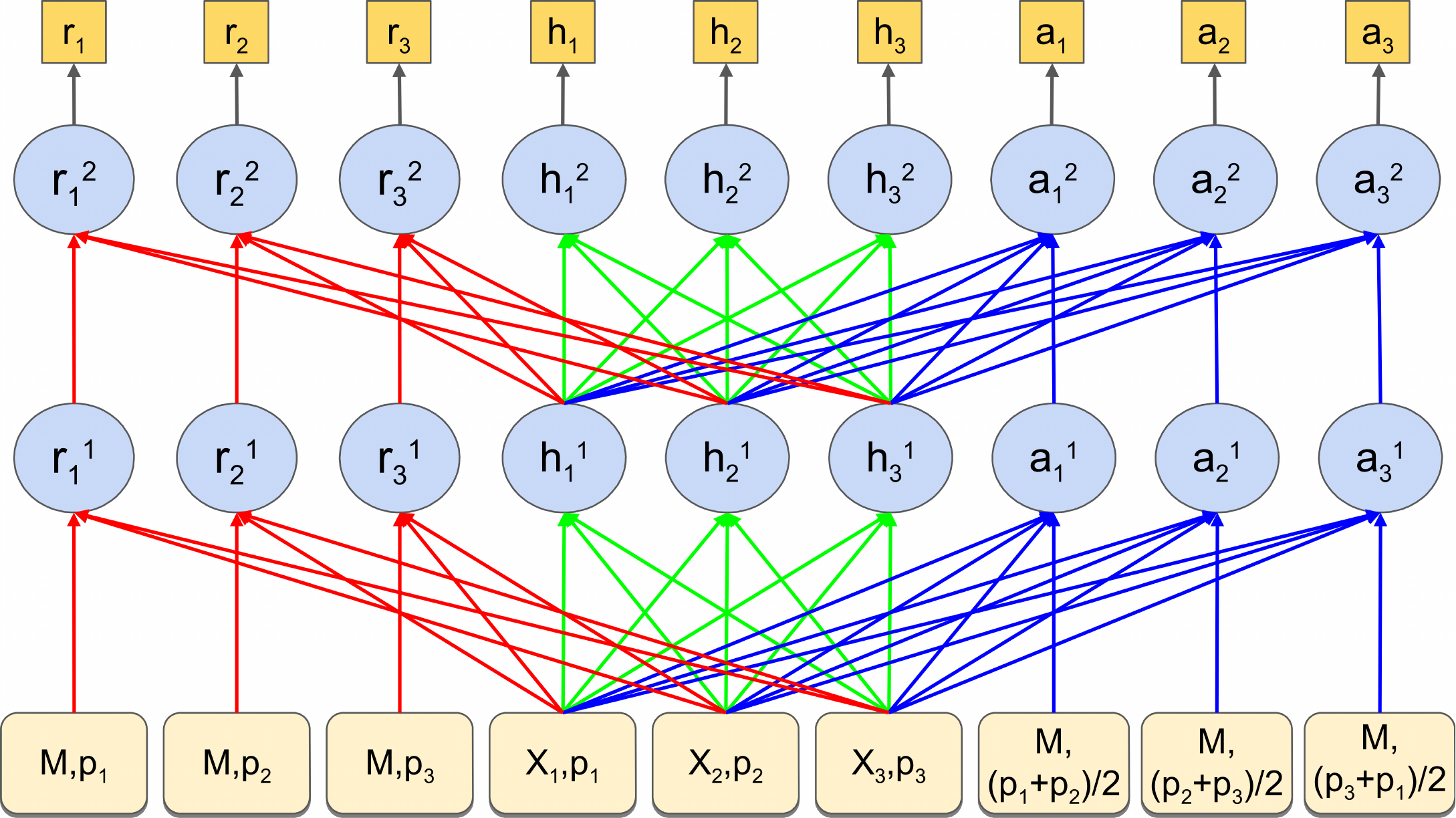}
    \caption{A Parallel Edit Architecture based on a 2-layer bidirectional transformer. Input sequence length is 3. Arrows indicate attention mask for computation of $\vh^l_i,\vr^l_i,\vi^l_i$ at position i for layer l.}
    \label{fig:pieArch}
\end{figure}

\subsubsection{An Edit-factorized BERT Architecture}
\label{sec:faclogic}
We adapt a pre-trained BERT-like bidirectional language model to learn to predict edits as follows.  For suitably capturing the contexts for replace edits, for each position $i$ we create an additional input comprising of $\vr^0_i = [\maskE, \vpos_i]$ where $\maskE$  is the embedding for $\maskT$ token 
in the LM.  Likewise for a potential insert between $i$ and $i+1$ we create an input  $\vi^0_i = [\maskE, \frac{\vpos_i+\vpos_{i+1}}{2}]$ where the second component is the average of the position embedding of the $i$th and $i+1$th position.  As shown in Figure~\ref{fig:pieArch}, at each layer $\ell$ we compute self-attention for the $i$th replace unit $\vr^\ell_i$ over $\vh_j^\ell$ for all $j \ne i$ and itself. 
Likewise, for the append unit $\vi^\ell_i$ the self-attention is over all $\vh^\ell_j$ for all $j$s and itself. 
At the last layer we have $\vh_1,\ldots,\vh_n,\vr_1,\ldots,\vr_n,\vi_1,\ldots,\vi_n$.  Using these we compute logits factorized over edits and their token argument.  For an edit $e \in {\cal E}$ at position $i$, let $w$ denote the argument of the edit (if any). As mentioned earlier, $w$ can be a q-gram for append and replace edits. Embedding of $w$, represented by $\phi(w)$ is obtained by summing up individual output embeddings of tokens in $w$. 
Additionally, in the outer layer we allocate edit-specific parameters $\vtheta$ corresponding to each distinct command in $\cE$.  
Using these, various edit logits are computed as follows: 
\begin{align*}
   & \Pr(e_i|\vx) = \text{softmax}(\text{logit}(e_i|\vx))~~~\text{where} \\
   & \text{logit}(e_i|\vx)= \numberthis \label{logit_decomposition}\\
    & \mkern-6mu \begin{cases} \theta_{\cpy}^{\intercal}\vu_i + \phi({x_i})^{\intercal}\vg_i + 0&\mkern-12mu\mbox{if~} e_i=\cpy\\
      \theta_{\app(w)}^{\intercal}\vu_i + \phi(x_i)^{\intercal}\vg_i + \phi(w)^{\intercal} \vi_i&\mkern-12mu\mbox{if~} e_i=\app(w)\\
  \theta_{\replace(w)}^{\intercal}\vu_i + 0 + (\phi(w) - \phi(x_i))^{\intercal}\vr_i&\mkern-12mu\mbox{if~} e_i=\replace(w)  \\
  \theta_{\delete}^{\intercal}\vu_i + 0 + 0&\mkern-12mu\mbox{if~} e_i=\delete \\
  \theta_{T_k}^{\intercal}\vu_i + \phi({x_i})^{\intercal}\vg_i + 0 &\mkern-12mu\mbox{if~} e_i=T_k \\
 \end{cases}
\end{align*}

The first term in the RHS of above equations captures edit specific score. 
The second term captures the score for copying the current word $x_i$ to the output. The third term models the influence of a new incoming token in the output obtained by a replace or append edit. For replace edits, score of the replaced word is subtracted from score of the incoming word. 
For transformation we add the copy score because they typically modify only the word forms, hence we do not expect meaning of the transformed word to change significantly.

The above equation provides insights on why predicting independent edits is easier than predicting independent tokens.  Consider the append edit  ($\app(w)$).  Instead of independently predicting $x_i$ at $i$ and $w$ at $i+1$, we jointly predict the tokens in these two slots and contrast it with not inserting any new $w$ after $x_i$ in a single softmax. We will show empirically (Sec~\ref{sec:ablation}) that such selective joint prediction is key to obtaining high accuracy in spite of parallel decoding.

Finally, loss for a training example ($\ve,\vx$) is obtained by summing up the cross-entropy associated with predicting edit $e_i$ at each token $x_i$. 
\begin{equation}
    \mathcal{L}(\ve,\vx) = - \Sigma_{i}\log(\Pr(e_i|\vx))
\end{equation}

\section{Experiments}
\label{Experiments}
We compare our parallel iterative edit (PIE) model with state-of-the-art GEC models that are all based on attentional encoder-decoder architectures. In Section~\ref{sec:others} we show that the PIE model is also effective on two other local sequence transduction tasks: spell and OCR corrections. Hyperparameters for all the experiments are provided in Table~\ref{tab:hparams_1},~\ref{tab:hparams_2} of the Appendix.
\subsection{Grammatical error correction (GEC)}\label{sec:gecexpts}
We use Lang-8~\cite{mizumoto2011mining}, NUCLE~\cite{ng2014conll} and FCE~\cite{yannakoudakis2011new} corpora, which jointly comprise 1.2 million sentence pairs in English.  The validation dataset comprises of 1381 sentence pairs from CoNLL-13 \cite{ngconll} test set.
We initialize our GEC model with the publicly available BERT-LARGE\footnote{\url{https://github.com/google-research/bert}} model that was pre-trained on Wikipedia (2500M words) and Book corpus \cite{zhu2015aligning} (800M words) to predict 15\% randomly masked words using its deep bidirectional context.  Next, we perform 2 epochs of training on a synthetically perturbed version of the One-Billion-word corpus~\cite{chelba2013one}. We refer to this as synthetic training. Details of how we create the synthetic corpus appear in Section~\ref{synthetic_gen} of the appendix.
Finally we fine-tune on the real GEC training corpus for 2 epochs. We use a batch size of 64 and learning rate 2e-5.\\
The edit space consists of copy, delete, $1000$ appends, $1000$ replaces and $29$ transformations and their inverse. Arguments of Append and Replace operations mostly comprise punctuations, articles, pronouns, prepositions, conjunctions and verbs. Transformations perform inflections like add suffix {\tt s, d, es, ing, ed} or replace suffix {\tt s to ing}, {\tt d to s}, etc. These transformations were chosen out of common replaces in the training data such that many replace edits map to only a few transformation edits, in order to help the model better generalize replaces across different words. In Section~\ref{sec:ablation} we see that transformations increase the model's recall. The complete list of transformations appears in Table~\ref{tab:suffix_transforms} of the Appendix. 
We evaluate on $F_{0.5}$ score over span-level corrections from the
MaxMatch (M2) scorer \cite{dahlmeier2012better} on CONLL-2014-test.
Like most existing GEC systems, we invoke a spell-checker on the test sentences before applying our model.
We also report $\text{GLEU}^+$ \cite{napoles2016gleu} scores on JFLEG corpus \cite{napoles-sakaguchi-tetreault:2017:EACLshort} to evaluate fluency. 
\begin{table}[H]
\centering
\begin{tabular}{|p{2.5cm}|c|c|c|c|}
\hline
Work & \multicolumn{3}{c|}{CONLL-14} & JFLEG \\ \cline{2-5} 
                      & P       & R       & $F_{0.5}$    & $\text{GLEU}^+$ \\ \hline
{\scriptsize \citet{zhao2019improving}}                  & 67.7    & 40.6    & 59.8      & -     \\ \hline
{\scriptsize \citet{lichtarge2019}}             & 65.5    & 37.1    & 56.8      & 61.6  \\ \hline
{\scriptsize \citet{chollampatt2018mlconv}}                  & 69.9    & 23.7    & 46.4      & 51.3  \\ \hline
PIE (This work)             & 66.1    & 43.0    & 59.7      & 60.3  \\ \hline
\end{tabular}
\caption{Comparison of non-ensemble model numbers for various GEC models. Best non-ensemble results are reported for each model.}
\label{tab:single_model_conll14}
\end{table}

\begin{table*}
\centering
\begin{tabular}{|p{2.5cm}|c|c|c|c|p{1.1cm}|}
\hline
Work            & Method                                                                                   & \multicolumn{3}{c|}{CONLL-14} & JFLEG \\ \cline{3-6} 
                                 &                                                                                                           & P        & R       & $F_{0.5}$   & $\text{GLEU}^+$  \\ \hline
{\scriptsize \citet{zhao2019improving}}     &  \begin{tabular}[c]{@{}c@{}}Transformer + Pre-training + LM \\ + Spellcheck + Ensemble Decoding (4)  \end{tabular}                                                     & 71.6       & 38.7    & 61.2     & 61.0    \\ \hline 
{\scriptsize \citet{lichtarge2019}}          & \begin{tabular}[c]{@{}c@{}}Transformer + Pre-training \\ + Iterative refinement + Ensemble Decoding (8) \end{tabular}                                                                                & 66.7     & 43.9    & 60.4     & 63.3  \\ \hline 
{\scriptsize \citet{Grundkiewicz2018}}                           & \begin{tabular}[c]{@{}c@{}}SMT + NMT (Bi-GRU) + LM  \\ + Ensemble Decoding (4) + Spellcheck\end{tabular} & 66.8    & 34.5   & 56.3    & 61.5 \\ \hline

{\scriptsize \citet{chollampatt2018mlconv}}     & CNN + LM + Ensemble Decoding (4) + Spellcheck                                                                                                      & 65.5     & 33.1    & 54.8     & 57.5  \\ \hline

PIE (This work) & \begin{tabular}[c]{@{}c@{}}Transformer + LM + Pre-training + Spellcheck \\ + Iterative refinement + Ensembles (5) \end{tabular}                                                               & 68.3    & 43.2   & 61.2    & 61.0  \\ \hline 
\end{tabular}
\caption{Comparison of recent GEC models trained using publicly available corpus. All the methods here except ours perform sequential decoding. Precision and Recall are represented by P and R respectively.}
\label{tab:ensemble_conll14}
\end{table*}
\subsubsection{Overall Results}
Table~\ref{tab:ensemble_conll14} compares ensemble model results of PIE and other state of the art models, which all happen to be seq2seq ED models and also use ensemble decoding. For PIE, we simply average the probability distribution over edits from 5 independent ensembles. In Table-\ref{tab:single_model_conll14} we compare non-ensemble numbers of PIE with the best available non-ensemble numbers of competing methods.
On CoNLL-14 test-set our results are very close to the highest reported by \citet{zhao2019improving}. These results show that our parallel prediction model is competitive without incurring the overheads of beam-search and slow decoding of sequential models.  $\text{GLEU}^+$ score, that rewards fluency, is somewhat lower for our model on the JFLEG test set because of parallel predictions. We do not finetune our model on the JFLEG dev set. We expect these to improve with re-ranking using a LM. 
All subsequent ablations and timing measurements are reported for non-ensemble models.

\subsubsection{Running Time Comparison}

\begin{figure}[h]
    \includegraphics[width=1.03\hsize,height=.8\hsize]{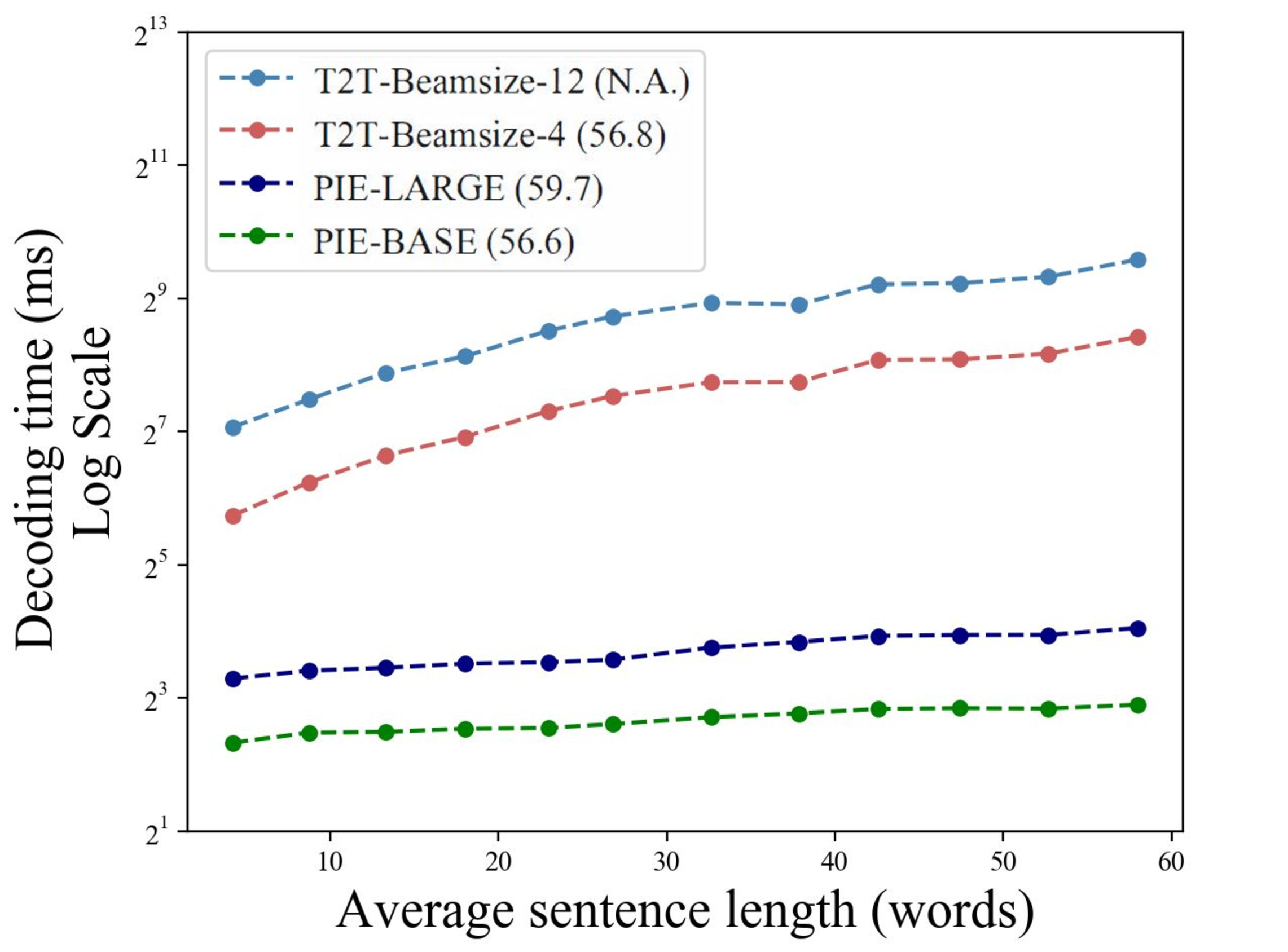}
    \caption{Comparing average decoding time in milliseconds (log scale) of PIE Vs transformer based ED model with two beam widths. Numbers in legend denote M2 score.} 
    \label{fig:decodeTime}
\end{figure}

Parallel decoding enables PIE models to be considerably faster than ED models. In Figure~\ref{fig:decodeTime} we compare wall-clock decoding time of PIE with 24 encoder layers (PIE-LARGE, $F_{0.5}=59.7$), PIE with 12 encoder layers (PIE-BASE, $F_{0.5}=56.6$) and competitive ED architecture by \citet{lichtarge2019} with 6 encoder and 6 decoder layers (T2T, $F_{0.5}=56.8$ ) on CoNLL-14 test set. All decoding experiments were run and measured on a n1-standard-2\footnote{\url{https://cloud.google.com/compute/docs/machine-types\#standard_machine_types}} VM instance with a single TPU shard (v-2.8). We observe that even PIE-LARGE is between a factor of 5 to 15 faster than an equivalent transformer-based ED model (T2T) with beam-size 4.  The running time of PIE-LARGE increases sub-linearly with sentence length whereas the ED model's decoding time increases linearly.

\subsubsection{Impact of Iterative Refinement}
\label{subsec:iter_ref_impact}
We next evaluate the impact of iterative refinements on accuracy in Table~\ref{tab:multiround_refinement}.
Out of 1312 sentences in the test set, only 832 sentences changed in the first round which were then fed to the second round where only 156 sentences changed, etc. The average number of refinement rounds per example was 2.7.  In contrast, a sequential model on this dataset would require 23.2 steps corresponding to the average number of tokens in a sentence. 
The $F_{0.5}$ score increases from 57.9 to 59.5 at the end of the second iteration.

\begin{table}[]
    \centering
    \begin{tabular}{|l|r|r|r|r|} \hline
    
         & R1 & R2 & R3 & R4  \\ \hline
        \#edited & 832 & 156 & 20 & 5  \\ \hline
        P      & 64.5 & 65.9 & 66.1 & 66.1   \\
        R      & 41.0 & 42.9 & 43.0 & 43.0   \\
        $F_{0.5}$ & 57.9 & 59.5 & 59.7 & 59.7  \\ \hline
    \end{tabular}
    \caption{Statistics on 
    successive rounds of iterative refinement. First row denotes number changed sentences (out of 1312) with each round on CONLL-14 (test).
    }
    \label{tab:multiround_refinement}
\end{table}{}

Table~\ref{tab:gec} presents some sentences corrected by PIE.  We see that PIE makes multiple parallel edits in a round if needed. Also, we see how refinement over successive iterations captures output space dependency. For example, in the second sentence {\tt interact} gets converted to {\tt interacted} followed by insertion of {\tt have} in the next round. 

\begin{table}[h]
\begin{small}
    \centering
    \begin{tabular}{|l|p{62mm}|} \hline

$\vx$ & I started invoving into Facebook one years ago .\\
PIE1 & I started {\color{red}\textbf{\textit{involving}}} {\color{red}\textbf{\textit{in}}} Facebook one {\color{red}\textbf{\textit{year}}} ago .\\
PIE2 & I started {\color{red}\textbf{\textit{involved}}} in Facebook one year ago .\\
PIE3 & I started {\color{red}\textbf{\textit{getting}}} involved in Facebook one year ago .\\ \hline
$\vx$ & Since ancient times , human interact with others face by face .\\
PIE1 & Since ancient times , humans {\color{red}\textbf{\textit{interacted}}} with others face {\color{red}\textbf{\textit{to}}} face .\\
PIE2 & Since ancient times , humans {\color{red}\textbf{\textit{have}}} interacted with others face to face .\\ \hline
$\vx$ & However , there are two sides of stories always .\\
PIE1 & However , there are {\color{red}\textbf{\textit{always}}} two sides {\color{red}\textbf{\textit{to}}} stories {\color{red}\st{always}} .\\
PIE2 & However , there are always two sides to {\color{red}\textbf{\textit{the}}} stories .\\
PIE3 & However , there are always two sides to the {\color{red}\textbf{\textit{story}}} .\\ \hline
    \end{tabular}
    \end{small}
    \caption{Parallel and Iterative edits done by PIE.}
    \label{tab:gec}
\end{table}{}

\setlength\tabcolsep{5.0pt}
\begin{table}[h]
    \centering
    \begin{tabular}{|r|l|r|r|r|} \hline
        \# & Methods & P & R & $F_{0.5}$ \\ \hline
        1& PIE   & 66.1 & 43.0 &  59.7 \\
        2&\hspace{0.5mm}$-$ Synthetic training & 67.2 & 34.2 & 56.3 \\
        3&\hspace{2.5mm}$-$Factorized-logits & 66.4 & 32.8 & 55.1  \\ 
        4&\hspace{2.5mm}$-$Append +Inserts & 57.4 & 42.5 & 53.6  \\
        5&\hspace{2.5mm}$-$Transformations & 63.6 & 27.9 & 50.6  \\
        6&\hspace{2.5mm}$-$LM Pre-init & 48.8 & 18.3  & 36.6 \\ 
      \hline
      7 & PIE on BERT-Base & 67.8 & 34.0  & 56.6  \\ \hline
    \end{tabular}
    \caption{Ablation study on the PIE model.}
    \label{tab:ablation}
\end{table}{}
\subsubsection{Ablation study on the PIE Architecture}
\label{sec:ablation}

In this section we perform ablation studies to understand the importance of individual features of the PIE model.

\noindent{\bf Synthetic Training}
We evaluate the impact of training on the artifically generated GEC corpus in row 2 of Table~\ref{tab:ablation}. We find that without it the $F_{0.5}$ score is 3.4 points lower.

\noindent{\bf Factorized Logits}
We evaluate the gains due to our edit-factorized BERT model (Section~\ref{sec:faclogic}) over the default BERT model (Section~\ref{sec:default}).
In Table~\ref{tab:ablation} (row 3) we show that compared to the factorized model (row 2) we get a 1.2 point drop in $F_{0.5}$ score in absence of factorization.

\noindent{\bf Inserts as Appends}
on the preceding word was another important design choice. 
The alternative of predicting {\tt insert} independently at each gap with a null token added to $\vocab_a$ performs
2.7 $F_{0.5}$ points poorly
(Table~\ref{tab:ablation} row 4 vs row 2).

\noindent{\bf Transformation edits} are significant as we observe a 6.3 drop in recall without them (row 5).

\noindent{\bf Impact of Language Model}
We evaluate the benefit of starting from BERT's pre-trained LM by reporting accuracy from an un-initialized network (row 6).  We observe a 20 points drop in $F_{0.5}$ establishing the importance of LMs in GEC.

\noindent{\bf Impact of Network Size}
We train the BERT-Base model with one-third fewer parameters than BERT-LARGE. From Table~\ref{tab:ablation} (row 7 vs 1) we see once again that size matters in deep learning! 

\subsection{More Sequence Transduction Tasks}
\label{sec:others}
We demonstrate the effectiveness of PIE model on two additional local sequence transduction tasks recently used in \cite{Ribeiro2018}.

\paragraph{Spell Correction}
We use the twitter spell correction dataset \cite{aramaki2010typo} which consists of 39172 pairs of original and corrected words obtained from twitter. 
We use the same train-dev-valid split as \cite{Ribeiro2018} (31172/4000/4000). We tokenize on characters, and our vocabulary $\vocab$ and $\vocab_a$ comprises the 26 lower cased letters of English. 

\paragraph{Correcting OCR errors}
We use the Finnish OCR data set\footnote{\url{https://github.com/mpsilfve/ocrpp}} by \cite{silfverberg2016data} comprising words extracted from Early Modern Finnish corpus of OCR processed newspaper text. 
We use the same train-dev-test splits as provided by \cite{silfverberg2016data}.
We tokenize on characters in the word.
For a particular split, our vocabulary $\vocab$ and $\vocab_a$ comprises of all the characters seen in the training data of the split. 

\paragraph{Architecture}
For all the tasks in this section, PIE is a 4 layer self-attention transformer with 200 hidden units, 400 intermediate units and 4 attention heads. No $\cL$ pre-initialization is done.  Also, number of iterations of refinements is set to 1.

\paragraph{Results}

\begin{table}[]
    \centering
    \begin{tabular}{|l|r|r|} \hline
         Methods & Spell & OCR \\ \hline
         Seq2Seq (Soft, lstm)& 46.3 & 79.9  \\ 
         Seq2Seq (Hard,lstm)  & 52.2 & 58.4  \\
         Seq2Seq (Soft-T2T) & 67.6 & 84.5 \\ \hline
         Ribeiro2018 (best model) & 54.1 & 81.8 \\ 
         PIE &  67.0 &  87.6 \\
        \hline
    \end{tabular}
    \caption{Comparing PIE with Seq2Seq models (top-part) and Ribeiro's parallel model on two other local sequence transduction tasks. 
    }
    \label{tab:edits}
\end{table}{}

\begin{table}[]
\centering
\begin{tabular}{|c|c|c|}
\hline
      & PIE   & Soft-T2T \\ \hline
Spell & 80.43 & 36.62     \\ \hline
OCR   & 65.44 & 43.02     \\ \hline
\end{tabular}
\caption{Wall-clock decoding speed in words/second of PIE and a comparable Seq2Seq T2T model.}
\label{tab:ng_decoding}
\end{table}

Table~\ref{tab:edits} presents whole-word 0/1 accuracy for these tasks on PIE and the following methods: \citet{Ribeiro2018}'s local transduction model (described in Section~\ref{sec:related}), and LSTM based ED models with hard monotonic attention \cite{AharoniG17} and soft-attention \cite{nmt} as reported in \cite{Ribeiro2018}. In addition, for a fair decoding time comparison, we also train a transformer-based ED model referred as Soft-T2T with 2 encoder, 2 decoder layers for spell-correction and 2 encoder, 1 decoder layer for OCR correction.
We observe that PIE's accuracy is comparable with ED models in both the tasks. Table~\ref{tab:ng_decoding} compares decoding speed of PIE with Soft-T2T in words/second.  
Since more than 90\% of words have fewer than 9 tokens and the token vocabulary $\Sigma$ is small, decoding speed-ups of PIE over ED model on these tasks is modest compared to GEC.

\section{Related Work}
\label{sec:related}
\paragraph{Grammar Error Correction (GEC)} is an extensively researched area in NLP. See \citet{ngconll} and \citet{ng2014conll} for past shared tasks on GEC, and this\footnote{\url{https://nlpprogress.com/english/grammatical\_error\_correction.html}} website
for current progress.
Approaches attempted so far include rules~\cite{felice2014grammatical}, classifiers~\cite{rozovskaya2016grammatical}, statistical machine translation (SMT)~\cite{junczys2016phrase}, neural ED models~\cite{chollampatt2018mlconv,junczys2018approaching,ge2018fluency}, and hybrids~\cite{Grundkiewicz2018}.  All recent neural approaches are sequential ED models that predict either word sequences~\cite{zhao2019improving,lichtarge2019} or character sequences~\cite{XieAAJN16} using either multi-layer RNNs~\cite{ji2017nested,Grundkiewicz2018} or CNNs\cite{chollampatt2018mlconv,ge2018fluency} or Transformers~\cite{junczys2018approaching,lichtarge2019}. 
Our sequence labeling formulation is similar to \cite{Yannakoudakis2017} and \cite{kaili2018simple} but the former
uses it to only detect errors
and the latter only corrects five error-types using separate classifiers.
Edits have been exploited in earlier GEC systems too but very unlike our method of re-architecting the core model to label input sequence with edits.
\citet{SchmaltzKRS17} interleave edit tags in target tokens but use seq2seq learning to predict the output sequence. \citet{chollampatt2018mlconv} use edits as features for rescoring seq2seq predictions. \citet{junczys2018approaching} use an edit-weighted MLE objective to emphasise corrective edits during seq2seq learning. 
\citet{stahlberg2019neural} use finite state transducers, whose state transitions denote possible edits, built from an unlabeled corpus to constrain the output of a neural beam decoder to a small GEC-feasible space.
\paragraph{Parallel decoding in neural machine translation}
\citet{kaiser18a} achieve partial parallelism by first generating latent variables sequentially to model dependency.
\citet{Stern2018} use a parallel generate-and-test method with modest speed-up.   \citet{gu2018} generate all tokens in parallel but initialize decoder states using latent fertility variables to determine number of replicas of an encoder state. We achieve the effect of fertility using delete and append edits. \citet{Lee2018} generate target sequences iteratively but require the target sequence length to be predicted at start. In contrast our in-place edit model allows target sequence length to change with appends.

\paragraph{Local Sequence Transduction}
is handled in \citet{Ribeiro2018} by first predicting insert slots in $\vx$ using learned  insertion patterns and then using a sequence labeling task to output tokens in $\vx$ or a special token {\tt delete}.   Instead, we output edit operations including word transformations. Their pattern-based insert pre-slotting is unlikely to work for more challenging tasks like GEC.
\citet{Koide2018} design a special edit-invariant neural network for being robust to small edit changes in input biological sequences.  This is a different task than ours of edit prediction. \citet{yin2018learning} is about neural representation of edits specifically for structured objects like source code. This is again a different problem than ours.

\section{Conclusion}
We presented a parallel iterative edit (PIE) model for local sequence transduction with a focus on the GEC task. 
Compared to the popular encoder-decoder models that perform sequential decoding, parallel decoding in the PIE model yields a factor of 5 to 15 reduction in decoding time.
The PIE model employs a number of ideas to match the accuracy of sequential models in spite of parallel decoding:  it predicts in-place edits using a carefully designed edit space, iteratively refines its own predictions, and effectively reuses state-of-the-art pre-trained bidrectional language models. In the future we plan to apply the PIE model to more ambitious transduction tasks like translation.

\paragraph{\bf Acknowledgements} This research was partly sponsored by a Google India AI/ML Research Award and Google PhD Fellowship in Machine Learning. We gratefully acknowledge Google's TFRC program for providing us Cloud-TPUs. We thank Varun Patil for helping us improve the speed of pre-processing and synthetic-data generation pipelines.

\bibliographystyle{acl_natbib}
\bibliography{ML}

\newpage
\onecolumn
\appendix

\section{Appendix}
\subsection{Modified Cost in Levenshtein Distance Algorithm}
We keep delete and insert cost as $1$ as usual, but for substitutions, we use $1 + \epsilon d$, where $d$ is the absolute difference between the number of characters of replaced and substituted word. We set $\epsilon$ to $0.001$.
 Table~\ref{tab:diff_example} shows two minimum edit diffs if the substitution penalty has no such offset. In Diff-1, \{\texttt{.}\} substitutes \{\texttt{,}\} and \texttt{Then} substitutes \texttt{then}, followed by insertion of \{\texttt{,}\}. In Diff-2, \{\texttt{.}\} is inserted after \texttt{sat} followed by \texttt{Then} substituting \{\texttt{,}\} , followed by \{\texttt{,}\} substituting \texttt{then} . In absence of offset in substitution penalty, both the diffs have edit distance of $3$. In presence of offset, Diff-1 has an edit-distance of $3$, while Diff-2 has an edit-distance of $3.006$, this allows Diff-1 to be preferred over Diff-2. As we observe, offset helps in selection of well aligned minimum edit diffs among multiple minimum edit diffs.

\newcommand{\wdel}{{w_{del}}}
\newcommand{\wins}{{w_{ins}}}
\newcommand{\wsub}{{w_{sub}}}
\algnewcommand{\LineComment}[1]{\State \(\triangleright\) #1}
\renewcommand{\algorithmicrequire}{\textbf{Input:}}
\renewcommand{\algorithmicensure}{\textbf{Output:}}

\begin{table}[H]
    \centering
\setlength\tabcolsep{1.0pt}
\begin{tabular}{l|cccccccccc} \hline
     $\vx$ &  \multicolumn{10}{|l}{~[ He sat , then he ran ]}\\ 
     $\vxo$ & \multicolumn{10}{|l}{  ~[ He sat . Then , he ran ]}\\
     op-1 & \multicolumn{10}{|l}{  ~[ $\cpy$ $\cpy$ $\subs({\tt ,} \;, {\tt .})$ $\subs({\tt then} \;, {\tt Then})$ $\ins({\tt ,})$ , $\cpy$ $\cpy$ ]}\\
     Diff-1 & \multicolumn{10}{|l}{~[ He sat -, +. -then +Then +, he ran ]} \\
     op-2 & \multicolumn{10}{|l}{  ~[ $\cpy$ $\cpy$ $\ins({\tt .})$ $\subs({\tt ,} \;, {\tt Then})$ $\subs({\tt then} \;, {\tt ,})$ , $\cpy$ $\cpy$ ]}\\
     Diff-2 & \multicolumn{10}{|l}{~[ He sat +. -, +Then -then +, he ran ]}\\
     
\hline 
\end{tabular}
\caption{Two diffs having same edit distance in the absence of offset in substitution penalty}

\label{tab:diff_example}
\end{table}

\subsection{Suffix transformations}

\begin{table}[h]
    \centering
    \begin{small}
    \begin{tabular}{|l|l|} \hline
        Transformation & Example \\ \hline
        \addsuffix($s$) &  play $\rightleftharpoons$ plays \\
        \addsuffix($d$) &  argue $\rightleftharpoons$ argued \\ 
        \addsuffix($es$) &  express $\rightleftharpoons$ expresses \\
        \addsuffix($ing$) &  play $\rightleftharpoons$ playing \\
        \addsuffix($ed$) &  play $\rightleftharpoons$ played\\
        \addsuffix($ly$) &  nice $\rightleftharpoons$ nicely \\
        \addsuffix($er$) &  play $\rightleftharpoons$ player\\
        \addsuffix($al$) &  renew $\rightleftharpoons$ renewal \\
        \addsuffix($n$) &  rise $\rightleftharpoons$ risen \\ 
        \addsuffix($y$) &  health $\rightleftharpoons$ healthy  \\
        \addsuffix($ation$) &  inform $\rightleftharpoons$ information  \\
        \transform{$e$}{$ing$} &  use $\rightleftharpoons$ using \\
        \transform{$d$}{$t$} &  spend $\rightleftharpoons$ spent\\ 
        \transform{$d$}{$s$} &  
        compared $\rightleftharpoons$ compares\\
        \transform{$s$}{$ing$} &  claims  $\rightleftharpoons$ claiming \\
        \transform{$n$}{$ing$} &  deafen $\rightleftharpoons$ deafening \\
        \transform{$nce$}{$t$} &  insistence $\rightleftharpoons$ insistent \\
        \transform{$s$}{$ed$} &  visits $\rightleftharpoons$ visited \\
        \transform{$ing$}{$ed$} &  using $\rightleftharpoons$ used \\
        \transform{$ing$}{$ion$} &  creating $\rightleftharpoons$ creation \\
        \transform{$ing$}{$ation$} &  adoring $\rightleftharpoons$ adoration \\
        \transform{$t$}{$ce$} &  reluctant $\rightleftharpoons$ reluctance \\
        \transform{$y$}{$ic$} &  homeopathy $\rightleftharpoons$ homeopathic \\
        \transform{$t$}{$s$} &  meant $\rightleftharpoons$ means \\
        \transform{$e$}{$al$} &  arrive $\rightleftharpoons$ arrival \\
        \transform{$y$}{$ily$} &  angry $\rightleftharpoons$ angrily \\
        \transform{$y$}{$ied$} &  copy $\rightleftharpoons$ copied \\
        \transform{$y$}{$ical$} &  biology $\rightleftharpoons$ biological \\
        \transform{$y$}{$ies$} &  family $\rightleftharpoons$ families \\\hline
    \end{tabular}
    \caption{29 suffix transformations and their corresponding inverse make total 58 suffix transformations. }
    \label{tab:suffix_transforms}
     \end{small}
\end{table}{}

\subsection{Artificial Error Generation}
\label{synthetic_gen}
\begin{figure}[h]
\begin{algorithmic}
\Require{U: dataset of clean sentences}
\State $AppendError \gets 0$
\State $VerbError \gets 1$
\State $ReplaceError \gets 2$
\State $DeleteError \gets 3$
\For{sentence in U}
    \State $errorCount\gets multinoulli(0.05,0.07,0.25,0.35,0.28)$
    \For{$i \in 1\dots errorCount $}
        \State $errorType \gets multinoulli(0.30, 0.25, 0.25, 0.20)$
        \State introduce error of type $errorType$
    \EndFor
\EndFor
\Return{}
\end{algorithmic}
\caption{Algorithm to introduce errors in clean dataset}
\label{fig:errorgen}
\end{figure}

Figure~\ref{fig:errorgen} shows the algorithm used to introduce artificial errors in clean dataset. Given a sentence, first the number of errors in that sentence is determined by sampling from a multinoulli (over $\{0\dots 4\}$). Similarly, an error is chosen independently from another multinoulli (over $\{AppendError, VerbError, ReplaceError, DeleteError \}$).
The distribution of the number of errors in a sentence and probability of each kind of error was obtained based on the available parallel corpus.  For append, replace and delete errors, a position is randomly chosen for the error occurrence. For append error the word in that position is dropped. For delete error a spurious word from a commonly deleted words dictionary is added to that position. For replace error, both the actions are done. For a verb error, a verb is chosen at random from the sentence and is replaced by a random verb form of the same word. Commonly deleted words are also obtained from the parallel corpus.

\subsection{Wall-clock Decoding Times}
\begin{table*}[h]
\begin{tabular}{|p{2.2cm}|p{0.75cm}|p{0.75cm}|p{0.75cm}|p{0.75cm}|p{0.75cm}|p{0.75cm}|p{0.75cm}|p{0.75cm}|p{0.75cm}|p{0.75cm}|p{0.75cm}|p{0.75cm}|}
\hline
\begin{tabular}[c]{@{}c@{}}Average\\  sentence\\  length (words) \end{tabular} & 4.30  & 8.76  & 13.30 & 18.00 & 22.96 & 27.79 & 32.64 & 37.85 & 42.54 & 47.41 & 52.6  & 58.8  \\ \hline
\begin{tabular}[c]{@{}c@{}}T2T-bs-4\\ (56.8)\end{tabular}             & 53.5  & 75.5  & 99.7  & 121.1 & 158.5 & 185.5 & 214.0 & 214.5 & 270.1 & 271.4 & 287.8 & 343.1 \\ \hline
\begin{tabular}[c]{@{}c@{}}T2T-bs-12 \\ (N.A.)\end{tabular}                                                            & 134.2 & 179.1 & 236.0 & 279.9 & 365.6 & 424.3 & 488.5 & 481.3 & 592.6 & 599.7 & 640.2 & 767.2 \\ \hline
\begin{tabular}[c]{@{}c@{}}PIE-BASE\\ (56.6)\end{tabular}            & 5.0   & 5.5   & 5.6   & 5.8   & 5.8   & 6.1   & 6.5   & 6.8   & 7.1   & 7.2   & 7.1   & 7.4   \\ \hline
\begin{tabular}[c]{@{}c@{}}PIE-LARGE\\ (59.7)\end{tabular}            & 9.8   & 10.6  & 10.9  & 11.4  & 11.6  & 11.9  & 13.5  & 14.3  & 15.2  & 15.4  & 15.4  & 16.6  \\ \hline
\end{tabular}
\caption{Wall clock decoding time in milliseconds for various GEC models}
\end{table*}

\newpage
\subsection{Hyperparameters}
\begin{table}[!hbt]
\centering

\begin{tabular}{|c|c|c|c|}
\hline
                         Hyperparameters       & \begin{tabular}[c]{@{}c@{}}PIE-BASE  GEC\end{tabular} & \begin{tabular}[c]{@{}c@{}}PIE-LARGE   GEC\end{tabular} & \begin{tabular}[c]{@{}c@{}}PIE  OCR/SPELL\\ Correction\end{tabular} \\ \hline
attention\_probs\_dropout\_prob & 0.1                                                     & 0.1                                                      & 0.1                                                                  \\ \hline
directionality                  & bi-directional                                                       & bi-directional                                                     & bi-directinoal                                                                    \\ \hline
hidden\_act                     & gelu                                                  & gelu                                                   & gelu                                                               \\ \hline
hidden\_dropout\_prob           & 0.1                                                     & 0.1                                                      & 0.1                                                                  \\ \hline
hidden\_size                    & 768                                                     & 1024                                                     & 200                                                                  \\ \hline
initializer\_range              & 0.02                                                    & 0.02                                                     & 0.02                                                                 \\ \hline
intermediate\_size              & 3072                                                    & 4096                                                     & 400                                                                  \\ \hline
max\_position\_embeddings       & 512                                                     & 512                                                      & 40                                                                   \\ \hline
num\_attention\_heads           & 12                                                      & 16                                                       & 4                                                                    \\ \hline
num\_hidden\_layers             & 12                                                      & 24                                                       & 4                                                                    \\ \hline
type\_vocab\_size               & 2                                                       & 2                                                        & 2                                                                    \\ \hline
vocab\_size                     & 28996                                                   & 28996                                                    & 110/26                                                                  \\ \hline
copy\_weight                    & 0.4     & 0.4 & 1 \\ \hline
\end{tabular}

\caption {Hyperparameters used in PIE Model for GEC, OCR Correction and Spell Correction. In GEC, copy weight of 0.4 is used (based on validation set) to scale down the loss corresponding to copy label for handling class imbalance}
\label{tab:hparams_1}

\vspace{1cm}

\centering

\begin{tabular}{|c|c|c|c|l}
\cline{1-4}
Hyperparameters           & T2T GEC                   & \begin{tabular}[c]{@{}c@{}}T2T OCR \\ Correction\end{tabular} & \begin{tabular}[c]{@{}c@{}}T2T Spell \\ Correction\end{tabular} &  \\ \cline{1-4}
T2T hparams set & transformer\_clean\_big\_tpu & transformer\_tiny                                            & transformer\_tiny                                               &  \\ \cline{1-4}
num\_encoder\_layers      & 6                         & 2                                                            & 2                                                               &  \\ \cline{1-4}
num\_decoder\_layers      & 6                         & 1                                                            & 2                                                               &  \\ \cline{1-4}
hidden\_size              & 1024                         & 200                                                          & 200                                                             &  \\ \cline{1-4}
filter\_size              & 4096                        & 400                                                          & 400                                                             &  \\ \cline{1-4}
\end{tabular}
\caption{Hyperparameters used in T2T transformer models for GEC, OCR Correction and Spell Correction. tensor2tensor (\url{https://github.com/tensorflow/tensor2tensor}) library was used for implementation}
\label{tab:hparams_2}
\end{table}
\end{document}